\pdfoutput=1
\documentclass[11pt]{article}

\usepackage[final]{acl}

\usepackage{times}
\usepackage{latexsym}
\usepackage{graphicx} 
\DeclareMathAlphabet\mathbfcal{OMS}{cmsy}{b}{n}
\usepackage{svg, svg-extract}
\usepackage{mathtools, amsfonts}
\usepackage[T1]{fontenc}

\usepackage[utf8]{inputenc}

\usepackage{microtype}
\usepackage{xurl}

\usepackage{inconsolata}

\usepackage{xcolor}

\title{LyS at SemEval-2024 Task 3: An Early Prototype for End-to-End Multimodal Emotion Linking as Graph-Based Parsing}

\author{Ana Ezquerro and David Vilares \\
  Universidade da Coruña, CITIC \\
  Departamento de Ciencias de la Computación y Tecnologías de la Información \\
  Campus de Elviña s/n, 15071 \\
  A Coruña, Spain\\
  \texttt{\{ana.ezquerro, david.vilares\}@udc.es}}
  
\DeclareFontFamily{OT1}{pzc}{}
\DeclareFontShape{OT1}{pzc}{m}{it}{<-> s * [1.10] pzcmi7t}{}
\DeclareMathAlphabet{\mathpzc}{OT1}{pzc}{m}{it}

\def\SPSB#1#2{\rlap{\textsuperscript{#1}}\SB{#2}}
\def\SP#1{\textsuperscript{#1}}
\def\SB#1{\textsubscript{#1}}

\definecolor{lightgreen}{HTML}{D9F2D0}
\definecolor{lightpink}{HTML}{F2CFEE}
\begin{document}
\maketitle
\begin{abstract}

This paper describes our participation in SemEval 2024 Task 3, which focused on Multimodal Emotion Cause Analysis in Conversations. We developed an early prototype for an end-to-end system that uses graph-based methods from dependency parsing to identify causal emotion relations in multi-party conversations. Our model comprises a neural transformer-based encoder for contextualizing multimodal conversation data and a graph-based decoder for generating the adjacency matrix scores of the causal graph. We ranked 7th out of 15 valid and official submissions for Subtask 1, using textual inputs only. We also discuss our participation in Subtask 2 during post-evaluation using multi-modal inputs.
\end{abstract}

\section{Introduction}

\begin{figure}[htbp!]
\centering
\includegraphics[width=0.47\textwidth]{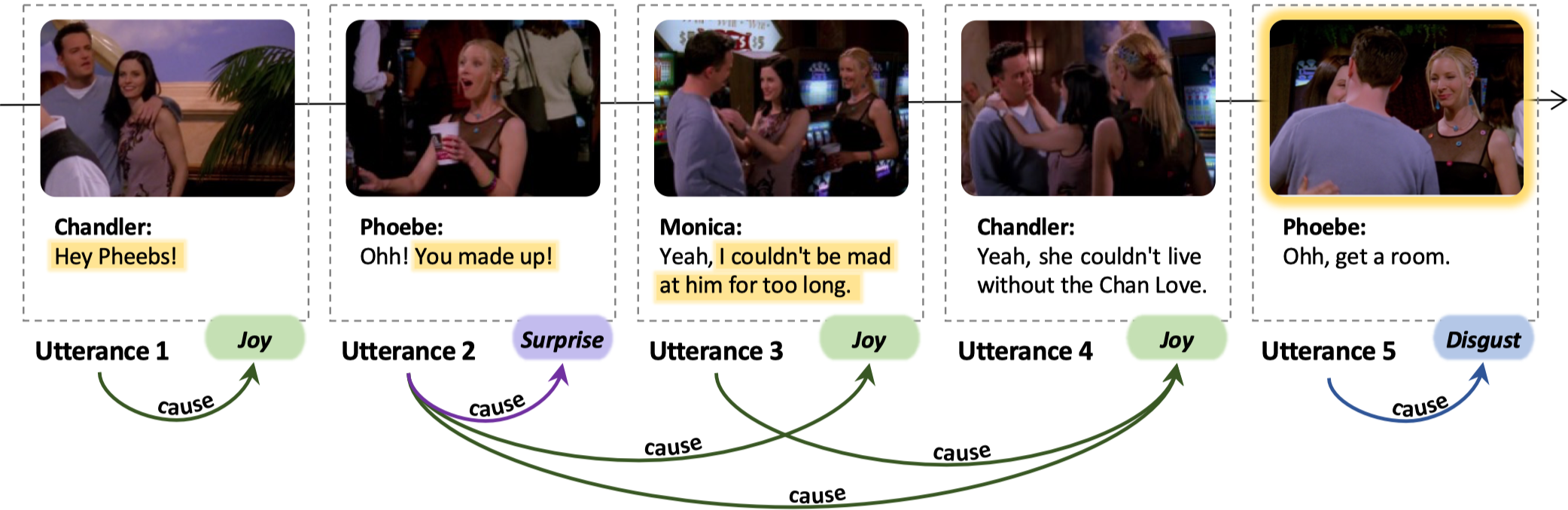} 
\caption{Example taken from the official website of the SemEval Task 3  - {\small\url{https://nustm.github.io/SemEval-2024_ECAC/}}. The goal of the task consists of predicting (i) the emotion associated to each utterance within the conversation, (ii) the cause-effect relations that trigger the emotions between utterances and (iii) the associated span in the cause utterance.}
\label{fig:example}
\end{figure}

SemEval 2024 Task 3 focused on Multimodal Emotion Cause Analysis in Conversations \cite{ECAC2024SemEval}. Figure \ref{fig:example} shows an example provided by the organizers to illustrate the task. Two subtasks were proposed: Subtask 1, which uses only textual inputs, and Subtask 2, which allows for the consideration of video and audio processing as well.

The shared task is timely given the recent success of multimodal architectures combining computer vision \cite{redmon2016look, wang2023internimage}, natural language processing \cite{devlin2019bert, beltagy2020longformer}, and speech processing \cite{gong2021ast, radford2022robust} advancements. In the particular context of multimodal emotion analysis, the task builds on top of previous work such as recognizing the triggered emotions as a classification task \cite{alhuzali2021spanemo, zheng2023facial} or predicting complex cause-effect relations between speakers \cite{wei2020effective, ding2020ecpe}.
For the particular case of the shared task, the dataset  - centered in English - relies on \cite{wang2023multimodal} and provides text, image and audio inputs. 

We had time and resources only to build a textual model for official participation in Subtask 1. We validated some multimodal baseline approaches using vision and audio inputs, but the computational resources required to fine-tune text and video data were beyond our range, so we participated in Subtask 2 only during post-evaluation. In what follows, we describe our approach. The implementation of our early prototype can be found at {\small\url{https://github.com/anaezquerro/semeval24-task3}}.

\paragraph{Contribution}

We propose an end-to-end multimodal prototype based on a large multimodal encoder to contextualize text, image and audio inputs with a graph-based decoder to model the cause-effect relations between triggered emotions within multi-party conversations. The large encoder joins pretrained architectures in text \cite{devlin2019bert}, image \cite{dosovitskiy2021image} and audio \cite{baevski2020wav2vec} modalities, while the decoder is adapted from the graph-based approaches in semantic parsing \cite{dozat2018simpler}. The model is trained end-to-end.

\section{Background} 
\paragraph{Multimodal Emotion Cause Analysis} 
A number of datasets collecting multi-party conversations \cite{poria2019meld, chatterjee2019semeval, firdaus2020meisd} have been published to train and test multimodal neural architectures. Simpler configurations involve recognizing the speaker emotion at each utterance - this task is commonly known as Emotion Recognition (ER) \cite{poria2019meld} - while others require a deeper level of understanding to model interactions and causal relations between speakers - Emotion-Cause Pair Extraction (ECPE) \cite{xia2019emotion}. The most common approaches follow an encoder-decoder neural architecture where the encoder is conformed by multiple modules - one module per input modality (text, image and/or audio) - and produces an inner representation at utterance level; and the decoder accepts the encoder outputs as inputs and returns a suitable output adapted to the specifications of the targeted task. In the context of Multimodal ER, \citet{nguyen2023conversation} proposed a GCN-based  decoder to capture temporal relations \cite{schlichtkrull2017modeling}, while \citet{dutta2024hcam} used cross-attention to fusion the input modalities and a final classification layer to predict the targeted emotions. Approaches in ECPE require an extra effort to represent and model causal information: \citet{wei2020effective} scored  all possible utterance tuples to predict the most probable list of emotion-cause pairs. Other authors, like \citet{chen2020end} and \citet{fan2020transition}, represented the emotion-cause pairs as a labeled graph between utterances and tried to predict the set of causal edges using a GCN or a transition-based system, respectively. The SemEval 2024 Task 3 joins the recognition and causal extraction tasks and challenges a system able to both model speaker emotions and elicit relations.

\paragraph{Graph-based decoding} 
For structured prediction tasks, such as dependency parsing, graph-based approaches are a standard for computing output syntactic representations \cite{mcdonald2006discriminative,martins-etal-2013-turning}. Particularly, \citet{dozat2017deep} introduced a classifier that computes a head and dependent representation for each token and then uses two biaffine classifiers: one computes a score for each pair of tokens to determine the most likely head, and the other determines the label for each head-dependent token pair. We will also build upon a biaffine graph-based parser: we will frame the task as predicting a dependency graph, where utterances are the nodes and emotions are dependency labels between pairs of utterances.

\section{System Overview}

\begin{figure}[h]
    \centering\small 
    \includegraphics[width=0.45\textwidth]{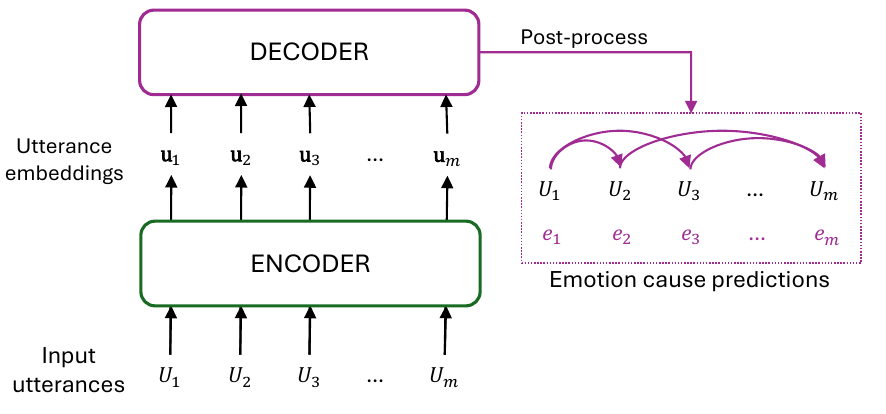}
    \caption{\label{fig:abstract} High-level architecture of our system. The encoder takes as input the sequence of $m$ utterances of a given conversation and returns a unique vector representation for each utterance. The decoder uses the utterance embedding matrix to apply the affine attention product in the decoder, obtain the scores of the adjacent matrix  and return the predicted sequence of emotions and the cause relations between utterances.}
\end{figure}

Our system consists of two modules: a large pretrained encoder  and a graph-based decoder (see Figure \ref{fig:abstract}). It can add extra input channels into the encoder without requiring any adjustments to the decoder, so the same decoder is used for both tasks, while the encoder is adapted to incorporate text-only (Subtask 1) or multimodal (Subtask 2) inputs.

Let $C=(U_1,...,U_m)$ be a conversation of $m$ utterances, where each utterance $U_i = \{W_i, s_i, \varepsilon_i\}$ is defined by (i) a sequence of  words $W_i=(w^{(i)}_1,...,w^{(i)}_{\ell|w,i|})$\footnote{From now on, we denote as $\ell|\cdot,i|$ the length of the $i$-th in a sequence $\cdot$, so $\ell|w,i|$ denotes the length of the $W_i$. Table \ref{tab:notation} summarizes the notation used in this paper.}, (ii) an active speaker $s_i$ and (iii) a triggered emotion $\varepsilon_i\in\mathcal{E}$\footnote{The set of emotions are described in \citet{wang2023multimodal}.}. The set of cause-pair relations between utterances can be represented as a directed labeled graph $G=(\mathcal{U}, \mathcal{R})$ where $\mathcal{U}=(U_1,...,U_m)$  is the sequence of utterances of the conversation assuming the role of the nodes of the graph and $\mathcal{R}=\{U_i\overset{\varepsilon_j}{\longrightarrow} U_j, \ i,j \in [1,m]\}$ is the set of emotion-cause  relations between an arbitrary cause utterance $U_i$ and its corresponding effect $U_j$. Thus, the task can be cast as the estimation of the adjacent matrix of $G$, similarly to syntactic \cite{ji2019graph} and semantic dependency \cite{dozat2018simpler} parsing. Adapting algorithms from parsing to model emotion-cause relations between utterances has also been explored by other authors, such as \citet{fan2020transition}, who instead explored a transition-based strategy.

\subsection{Textual Extraction}\label{subsec:subtask1}
The first subtask draws from only textual information to predict the adjacent matrix of $G$ with a span that covers the specific words from $U_i$ that trigger the emotion $\varepsilon_j$ in the cause relation  $U_i\overset{\varepsilon_j}{\longrightarrow} U_j$. 

\begin{figure}[h]
    \small 
    \includegraphics[width=0.48\textwidth]{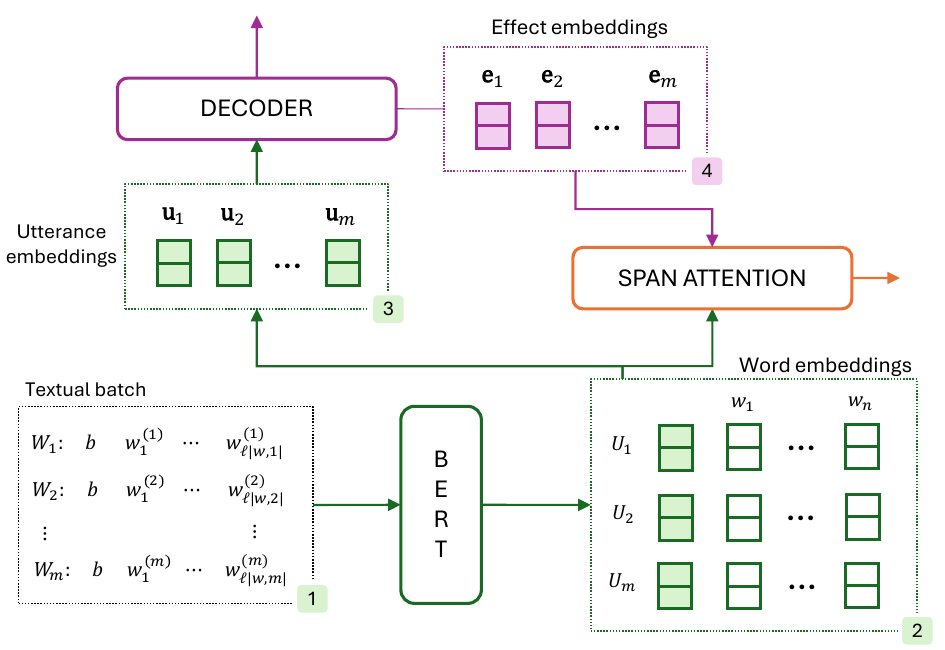}
    \caption{\label{fig:encoder1}High level representation of the textual encoder. The input (\colorbox{lightgreen}{\scriptsize1})
 is the matrix of stacked token vectors of each utterance. The last hidden states of BERT are used as word embeddings (\colorbox{lightgreen}{\scriptsize2}) and the special CLS tokens are used as utterance embeddings (\colorbox{lightgreen}{\scriptsize3}). The effect embeddings (\colorbox{lightpink}{\scriptsize4}) - a partial representation from the decoder - are taken as input to the span module with the contextualized BERT embeddings.}
\end{figure}

\paragraph{Textual encoder} 
Figure \ref{fig:encoder1} illustrates our encoder. Given the sequence of utterances ($U_1,...,U_m$), we encoded with BERT the batched sequence of utterances where each word sequence was preceded by the  CLS token \cite{devlin2019bert}. For each $U_i$, we select the CLS embedding ($\mathbf{u}_i$) from the contextualized embedding matrix $\mathbf{W}_i=(\mathbf{u}_i, \mathbf{w}_1,...,\mathbf{w}_{\ell|w,i|})$, which is assumed to have information of the whole sentence. The CLS embedding matrix $\mathbf{U}=(\mathbf{u}_1,...,\mathbf{u}_m)$ was passed as  input to the decoder module and the word embeddings were reserved for the span attention module.

\begin{figure}\small 
    \includegraphics[width=0.48\textwidth]{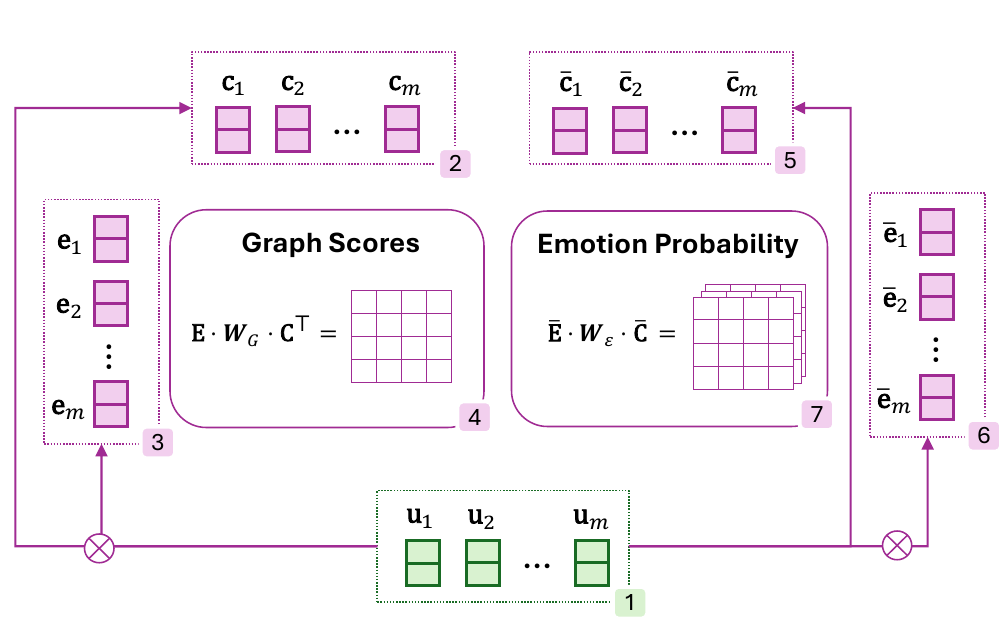}
    \caption{\label{fig:decoder1}Graph-based decoder. The utterance embeddings (\colorbox{lightgreen}{\scriptsize1}) are projected to different representations (\colorbox{lightpink}{\scriptsize2}, \colorbox{lightpink}{\scriptsize3}, \colorbox{lightpink}{\scriptsize5} \colorbox{lightpink}{\scriptsize 6}) using four feed-forward networks to flexibly represent utterance embeddings. The scores of the adjacent matrix and the probability tensor are computed with the affine attention product.}
\end{figure}

\paragraph{Graph-based decoder} 
Figure \ref{fig:decoder1} shows the forward-pass of the graph-based decoder from the encoder output of Figure \ref{fig:encoder1}. To produce an adjacent matrix $\mathbf{G}$ of dimensions\ $m\times m$, where each position $(i,j)$ represents the probability of a causal relation from $U_i$ (cause) to $U_j$ (effect), the first biaffine module uses a trainable matrix $\boldsymbol{\mathcal{W}}_\text{G}\in\mathbb{R}^{d_\text{G}\times d_\text{G}}$ and maps $\mathbf{U}$ using two feed-forward networks to a cause ($\mathbf{C}$) and  an effect ($\mathbf{E}$) representation. By projecting the original BERT embeddings to two different representations, $\mathbf{u}_i \sim (\mathbf{c}_i, \mathbf{e}_i)$, the decoder learns different contributions for the same utterance depending on the role. The affine product is defined as $\mathbf{G} =\mathbf{E}\cdot \boldsymbol{\mathcal{W}} \cdot \mathbf{C}^\top$. The second biaffine module uses a trainable tensor $\boldsymbol{\mathcal{W}}_\varepsilon\in\mathbb{R}^{d_\text{G}\times |\mathcal{E}| \times d_\text{G}}$ to predict the probabilities of triggered emotions between cause-effect utterances.

\begin{figure}
    \centering\small 
    \includegraphics[width=0.49\textwidth]{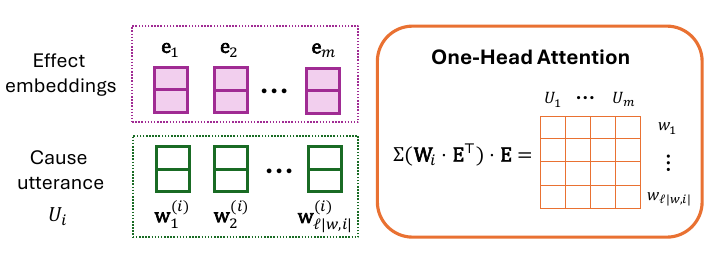}
    \caption{\label{fig:span-decoder}Span Attention module adapted from \citet{vaswani2017attention}. The tensor of word embeddings ($\mathbf{W}_1\cdots \mathbf{W}_m$) from the encoder (Figure \ref{fig:encoder1}) and the effect contextualizations ($\mathbf{E}$) from the decoder (Figure \ref{fig:decoder1}) are passed to the attention product using each $\mathbf{W}_i$ as \textit{key} and  $\mathbf{E}$ as \textit{query} and \textit{value} matrices.}
\end{figure}

\paragraph{Span Attention module}
To maintain the end-to-end prediction while learning the span associated to each relation $U_i\to U_j$, we created a binary tensor $\mathbf{S}=(\mathbf{S}_1\cdots \mathbf{S}_m)$ of dimensions $m \times m \times \max_{i=1,...,m}\{\ell|w,i|\}$\footnote{Note that each matrix $\mathbf{S}_i$ has dimensions $m\times \max_{i=1,...,m}\{\ell|w,i|\}$ and is associated to a \emph{cause} utterance.} to specify if a word $w_k\in W_i$ of $U_i$ is included in the span that triggers an emotion in $U_j$. To compute each $\mathbf{S}_i$, the matrix of word embeddings ($\mathbf{W}_i$)  of the utterance $U_i$ is passed through a One-Head Attention module (see Figure \ref{fig:span-decoder}), where $\mathbf{W}_i$ acts as the query matrix and $\mathbf{E}$ as the key and value matrices, so $\mathbf{S}_i = \Phi(\text{softmax}(\mathbf{W}_i\cdot  \mathbf{E}^\top)\cdot  \mathbf{E})$, where $\Phi$ is a feed-forward network to project the embedding dimension to a unique binary value.

\paragraph{Encoding speaker information}
The dataset includes information about the active speakers in each utterance. A first approach to use this information as input would be concatenating the speaker embeddings to the sequence of utterances. However, this might lead to some issues:  the model could assume that there is some inner dependency between triggered emotions and the characters in the conversation. This might be true in some cases, but it can also lead to biases, and there is still the challenge of modeling infrequent and unknown characters. To deal with this, we encoded a conversation $C$ with speakers $s_1,...,s_m$ using relative positional embeddings. For instance, the sequence (Chandler, Phoebe, Monica, Chandler, Phoebe) in Figure \ref{fig:example} would be encoded as $(0, 1, 2, 0, 1)$.

\subsection{Multimodal Extraction}\label{subsec:subtask2}

The second subtask adds a short video representation to each utterance, so $U_i$ in a conversation $C=(U_1,...,U_m)$ is now a tuple of five different elements $U_i=\{ W_i, s_i, \varepsilon_i, \mathbf{X}_i, \mathbf{a}_i\}$. The last two added items encode the image and audio: (i) $\mathbf{X}_i=(\mathbf{x}^{(i)}_1,...,\mathbf{x}^{(i)}_{\ell|x,i|})$ is the sequence of frames of the input video,  where each frame is an image\footnote{All frames are RGB images, being the majority resolution {\scriptsize$720\times 1280$}.} tensor of dimensions $h\times w\times 3$ and (ii)   $\mathbf{a}_i$ is the sampled audio signal of arbitrary length.

\paragraph{Image encoding} 
We relied on a Transformer-based architecture \cite{ma2022facial, zheng2023facial} to contextualize input images.  While recent studies have proposed adaptations of the Vision Transformer and 3-dimensional convolutions that capture temporal correlations between sequences of frames for video classification \cite{arnab2021vivit, ni2022expanding}, our experiments were constrained by our resource limitations, preventing us from using these pretrained architectures. Hence, for our multimodal baseline we opted for the the smallest version of the Vision Transformer (ViT) model \cite{dosovitskiy2021image} pretrained on the Facial Emotion Recognition dataset \cite{goodfellow2013challenges}\footnote{\scriptsize\url{https://huggingface.co/trpakov/vit-face-expression}.} to contextualize a small fraction of sampled frames\footnote{For our experiments we used 5 interleaved frames per video, although a lower sampling rate can be considered depending on the computational capabilities.}, and incorporated an LSTM-based module to derive a unique image representation for each utterance. From an image batch $\mathbf{X}_i$, each image $\mathbf{x}^{(i)}_k\in\mathbb{R}^{h\times w \times 3}$ was passed to the ViT base model to recover the output of the last hidden layer  and introduce it as input to the LSTM module to recover a final representation for $U_i$.

\paragraph{Audio encoding}
For our multimodal system we used the hidden contextualizations of the base version of wav2vec 2.0 \cite{baevski2020wav2vec}\footnote{\scriptsize\url{https://huggingface.co/facebook/wav2vec2-base-960h}.}. Given a raw audio ($\mathbf{a}_i$) of an utterance $U_i$, the encoder of wav2vec 2.0 returns a sequence of hidden states that we summarized with an additional trainable LSTM layer to retrieve a unique vector that contents the audio information.

\begin{figure}[h]
    \centering\small 
    \includegraphics[width=0.5\textwidth]{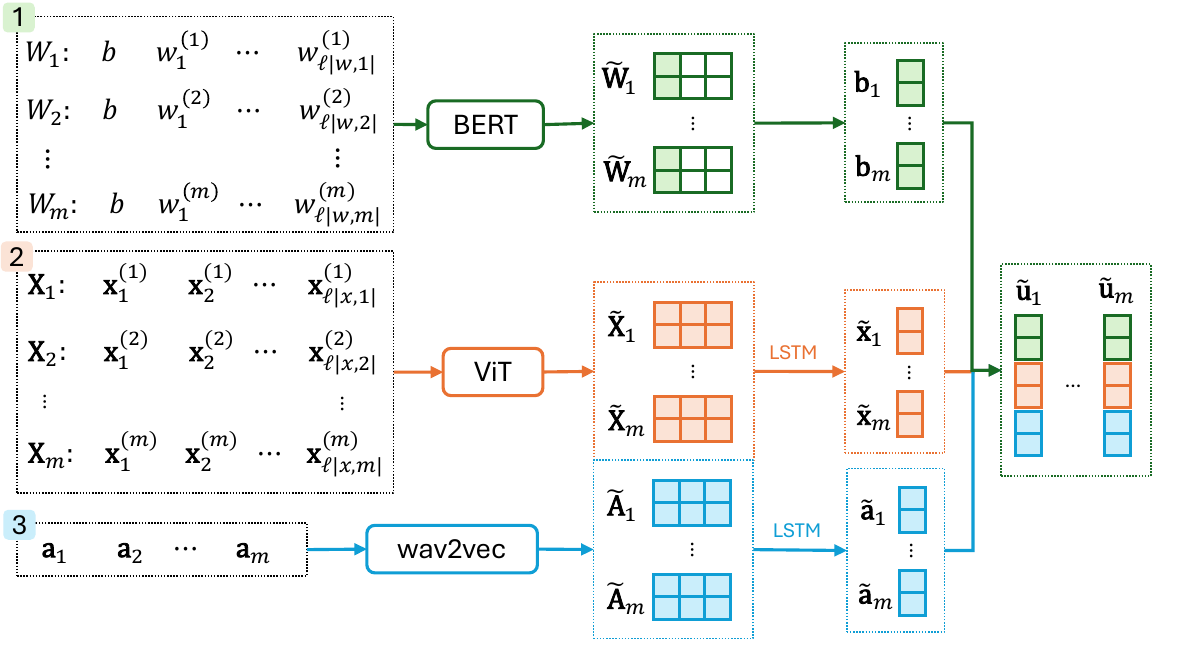}
    \caption{\label{fig:encoder2}Multimodal encoder for Subtask 2.}
\end{figure}
\paragraph{Model fine-tuning} The multimodal encoder (\S\ref{subsec:subtask2}) uses three pretrained architectures to contextualize individual utterances and passes to the decoder the concatenation of the three unimodal representations (Figure \ref{fig:encoder2}).
We chose to fine-tune only BERT during training together with the rest of the network. This was based on our empirical observation of superior results when learning from text compared to image and audio data. We entrusted the learning of audiovisual data to the LSTM learnable module within the encoder, presuming an accurate initial  contextualization from wav2vec 2.0 and ViT pretrained on FER-2013.

\subsection{Post-processing} Our end-to-end system directly recovers the predicted emotion-cause relations in a single post-processing step that linearly operates with the output tensors of the decoder. For the first subtask, the decoder returns (i) the adjacent matrix $\mathbf{G}\in\mathbb{R}^{m\times m}$, (ii) the labeled adjacent matrix $\overline{\mathbf{G}} \in\mathbb{R}^{m\times m \times |\mathcal{E}|}$ and (iii) the span scores $\mathbf{S}\in\mathbb{R}^{m\times m \times \ell_{\max} |w,i|}$. As \citet{dozat2017deep}, each arc $U_i\to U_j$ is predicted by thresholding $\mathbf{G}$, and, once the arcs are predicted, the tensor $\overline{\mathbf{G}}$ determines the label (emotion) associated to each arc. Since our formalization (\S\ref{subsec:subtask1}) associates a given utterance to an unique emotion, we leveraged the scores of $\overline{\mathbf{G}}$ by the cause utterances and return the emotion with highest score. Finally, to produce a continuous span for each score vector $\mathbf{s_{ij}}$, we considered the leftmost and rightmost elements of $\mathbf{s}_{ij}$ higher than a fixed threshold.

\begin{table}[htpb!]
    \centering\small 
    \renewcommand{\arraystretch}{1.5}
    \setlength{\tabcolsep}{5pt}
    \begin{tabular}{c|ccc|ccc}
        \hline 
         \textbf{\textsc{{ST-1}}} & \textbf{P}\SPSB{w}{s} &  \textbf{R}\SPSB{w}{s} & \SP{$\dagger$}\textbf{F}\SPSB{w}{s}  &  \textbf{P}\SPSB{w}{p} & \textbf{R}\SPSB{w}{p} & \textbf{F}\SPSB{w}{p} \\
         \hline 
         \textbf{BERT}\SB{400} & 10.19 & 5.46 &  7.01 & 21.64 & 15.09 & 17.33   \\
         \textbf{BERT}\SB{600} & 12.61 & 7.43 & 9.32 &  22.06 & 15.2 & 17.95  \\
         \textbf{BERT}\SB{800}  & 14.89 & 7.36 & 9.75 &  22.13 &  23.25 & 15.32  \\
         \hline 
    \end{tabular}
    \\[1em]
    \begin{tabular}{c|ccc}
        \hline 
          \textbf{\textsc{ST-2}}& \textbf{P}\SP{w} & \textbf{R}\SP{w} & \SP{$\dagger$}\textbf{F}\SP{w} \\
          \hline 
          \textbf{BERT} & 27.49, & 17.62, & 20.43 \\ 
          \textbf{+ViT} &  22.38 & 22.72 & 22.17   \\ 
          \textbf{+w2v}  &  28.4 & 20.01 & 23.36\\
           \textbf{+w2v+ViT} &    23.37 & 7.62 & 11.49  \\ 
           \hline 
    \end{tabular}
    \caption{\label{tab:results} Evaluation of our prototype with different multimodal configurations. Precision (P), recall (R) and F-Score (F) measured the weighted average across the eight emotions of the dataset (superscript \textbf{w} denotes that the measure is weighted) and for the first subtask the span performance is considered with strict correctness (subscript \textbf{s}) or overlapping (subscript \textbf{p}). The symbol $\dagger$ remarks the reference metric for each subtask.}
\end{table}

\section{Experiments}

\paragraph{Validation} 
The annotated dataset contains 1\,375 multi-party conversations with a total of 13\,619 utterances \cite{wang2023multimodal}. Although an unbiased estimation of the performance of our system would require validating the trained architecture using all available annotated data,  our time and resources limitations prevented us from conducting k-fold cross-validation. Instead, we partitioned  a 15\% of the annotated dataset as our development set. The specific split used will be available with the accompanying code to replicate our findings.

\paragraph{Evaluation} We use the official metrics\footnote{\scriptsize\url{https://github.com/NUSTM/SemEval-2024_ECAC/tree/main/CodaLab/evaluation}.}: the weighted strict F-Score for the Subtask 1 and the weighted F-Score for the Subtask 2. 

\paragraph{Hyperparameter configuration} Our computational limitations prevented us from exhaustively searching the optimal hyperparameters for our system. We conducted some tests varying the pretrained text encoder\footnote{We performed some experiments using all the versions of BERT, \cite{devlin2019bert}, RoBERTa \cite{liu2019roberta} and XLM-RoBERTa \cite{conneau2020unsupervised} and selected the best-performing textual encoder (BERT-large).},  model dimension, gradient descent algorithm and learning rate and adding or removing the speaker module. We maintained in all experiments some regularization techniques (such as dropout in the hidden layers and gradient norm clipping) to avoid over-fitting. Our final configuration uses AdamW optimizer \cite{loshchilov2019decoupled} with learning rate of $10^{-6}$ and is trained during one hundred epochs with early stopping on the validation set.

\section{Results}

Table \ref{tab:results} presents the performance of our system for both subtasks. For the first subtask, we investigated various embedding sizes of the Biaffine decoder while concurrently fine-tuning the largest version of BERT\footnote{\scriptsize\url{https://huggingface.co/google-bert/bert-large-cased}}. For the second subtask, we conducted experiments using different types of inputs to evaluate their impact. These included: (i) using only text-based inputs, (ii) adding audio data, (iii) incorporating visual data through frames, and (iv) leveraging all available multimodal inputs together. For approaches (i), (ii) and (iii), only BERT was fine-tuned, whereas for approach (iv), all pretrained weights were frozen. These weights solely served to contextualize input information, with the learning process confined to the decoder component.

Our top-performing model for the first subtask achieved a validation score of 9.75 and ranked in the evaluation set in 7th position among 15 participants with 6.77 points. We observed a slight performance improvement by increasing the hidden dimension of the decoder. 
Thus, considering the expansion of decoder layers could improve the performance. It is worth noting the significant impact of span prediction on the model performance: the proportional results consistently outperform strict metrics. Removing span prediction while retaining only text inputs results in a notable increase in F-Score (20.43 points for the second subtask), indicating the crucial role of span prediction in model learning. 
Furthermore, we noticed that there was a consistent delay in the alignment between recall and precision metrics, with precision consistently exceeding recall by more than 5 points across all approaches. This suggests that our system tends to adopt a conservative behavior, avoiding the number of false cause emotion predictions.

The best validation performance for the second subtask is achieved through the integration of text and audio, yielding a score of 23.36 points in the weighted F-Score. Using image data also improves the text-only baseline, though unexpectedly lags behind the audio model. It is important to note that these two approaches are not directly comparable due to differences in their data inputs: the text and image model only considers a fixed number of sampled frames, suggesting that providing more image data (ideally, the full sequence of frames) could potentially yield a better performance that surpasses the audio-based approach.  Unfortunately, we could not fine-tune BERT with the full multimodal encoder, so we were restricted to projecting the multimodal inputs to their respective contextualizations, and relying on the trainable weights of the decoder to optimize the full architecture. The results prove the importance of, at least, fine-tuning the text encoder: the F-Score only reaches 11.25 points, whereas the text finetuned baseline nearly doubles its performance with 20.43 points, highlighting the insufficient context of the original pretrained BERT embeddings to address this task. 

Once the post-evaluation period concluded, we upload an experimental submission of our best multimodal system to the official competition. We obtained 15.32 points in the weighted F-Score, positioning our baseline in the 13th place out of 18 participants.

\section{Conclusion}

We proposed a graph-based prototype for the analysis emotion-cause analysis in conversations. Given the limited preparation time, we only submitted official results for Subtask 1 (text-only), but also report post-evaluation results for Subtask 2 (multimodal). The task required predicting several aspects of the conversation: (i) the emotion associated with each utterance, (ii) the cause-effect relationships triggering these emotions between utterances, and (iii) the specific span within the cause utterance responsible for the emotion. We achieved 7th place out of 15 valid submissions for Subtask 1, a promising outcome considering the time and resource constraints we had to prepare the task. Yet, our results make us optimistic about exploring future research avenues to enhance our system and study lighter approaches  that can perform competitively. As future work, we aim to experiment with smaller and distilled models to encode textual, visual, and audio inputs, enabling us to fine-tune the full model cheaply.

\section*{Acknowledgments}
This work has received supported by Grant GAP (PID2022-139308OA-I00) funded by MCIN/AEI/10.13039/501100011033/ and by ERDF, EU; the European Research Council (ERC), under the Horizon Europe research and innovation programme (SALSA, grant agreement No 101100615); Grant SCANNER-UDC (PID2020-113230RB-C21) funded by MICIU/AEI/10.13039/501100011033; Xunta de Galicia (ED431C 2020/11); by Ministry for Digital Transformation and Civil Service and ``NextGenerationEU''/PRTR under grant TSI-100925-2023-1; and Centro de Investigación de Galicia ‘‘CITIC’’, funded by the Xunta de Galicia through the collaboration agreement between the Consellería de Cultura, Educación, Formación Profesional e Universidades and the Galician universities for the reinforcement of the research centres of the Galician University System (CIGUS).

\bibliography{custom}

\appendix

\section{Appendix}

\begin{table}[h]
    \scriptsize 
    \setlength{\tabcolsep}{2pt}
    \renewcommand{\arraystretch}{1.3}
    \begin{tabular}{c|p{6.7cm}}
        \hline 
         \textbf{I}&\textbf{Description} \\
         \hline 
         $U_i$ & Utterance $i$, defined as $U_i=(W_i, s_i, \varepsilon_i, \mathbf{X}_i, \mathbf{a}_i)$. \\
         $W_i$ & Word sequence of $U_i$ as $W_i=(w_1,...,w_{\ell|w,i|})$ \\ 
         $s_i$ & Speaker of $U_i$, where $s_i\in\mathcal{S}$ \\ 
         $\varepsilon_i$ & Emotion triggred in $U_i$, where $\varepsilon_i\in\mathcal{E}$ \\ 
         $\mathcal{S}$ & Set of speakers in the dataset.\\
         $\mathcal{E}$ & Set of annotated emotions.\\
         $\mathbf{X}_i$ & Sequence of frames of $U_i$ as $\mathbf{X}_i =(\mathbf{x}_1,...,\mathbf{x}_{\ell|x,i|})$\\ 
         $\mathbf{x}^{(i)}_k$ & Specific frame of $\mathbf{X}_i$, where $\mathbf{x}_k^{(i)}\in\mathbb{R}^{h\times w \times 3}$. \\ 
         $\mathbf{a}_i$ & Sampled audio signal of $U_i$, where $\mathbf{a}_i\in\mathbb{R}^{\ell|a,i|}$. \\ 
         $\ell|w,i|$ & Length of the sequence $W_i$. \\ 
         $\ell|x,i|$ & Lenght of the sequence $\mathbf{X}_i$\\ 
         \hline 
         \textbf{E} & \textbf{Description} \\
         \hline
         $\mathbf{u}_i$ & Encoder hidden representation of $U_i$ from BERT, where $\mathbf{u}_i \in\mathbb{R}^{1024}$. \\ 
         $\mathbf{W}_i$ & BERT word embeddings of $W_i$ as $\mathbf{W}_i=(\mathbf{u}_i,\mathbf{w}_1^{(i)},...,\mathbf{w}_{\ell|w,i|}^{(i)}).$ \\ 
         $\tilde{\mathbf{x}}_i$ & Visual hidden representation for $U_i$, obtained as $\tilde{\mathbf{x}}_i = \text{LSTM}_\text{x}^{-1}(\text{ViT}(\mathbf{X}_i))\in\mathbb{R}^{d_\text{V}}$. \\ 
         $\tilde{\mathbf{a}}_i$ & Audio hidden representation for $U_i$, obtained as $\tilde{\mathbf{a}}_i=\text{LSTM}_\text{a}^{-1}(\text{wav2vec}(\mathbf{a}_i))\in\mathbb{R}^{d_\text{a}}$. \\ 
         $\tilde{\mathbf{u}}_i$ & Multimodal representation for $U_i$ as $\tilde{\mathbf{u}}_i=(\mathbf{u}_i | \tilde{\mathbf{x}}_i |\tilde{\mathbf{a}}_i)$. \\ 
         \hline 
         \textbf{D} & \textbf{Description} \\ 
         \hline 
         $\Phi$ & Arbitrary feed-forward network.\\
         $\mathbf{c}_i$ & Cause embedding for $U_i$ as $\mathbf{c}_i = \Phi_c(\mathbf{u}_i)\in\mathbb{R}^{d_\text{G}}$. \\ 
         $\mathbf{e}_i$ & Effect embedding for $U_i$ as $\mathbf{e}_i=\Phi_e(\mathbf{u}_i)\in\mathbb{R}^{d_\text{G}}$. \\ 
         $\overline{\mathbf{c}}_i$ & Emotion cause embedding for $U_i$ as $\overline{\mathbf{c}}_i = \Phi_{c,\varepsilon}(\mathbf{u}_i)\in\mathbb{R}^{d_\text{G}}$. \\ 
         $\overline{\mathbf{e}}_i$ & Emotion effect embedding for $U_i$ as $\overline{\mathbf{e}}_i=\Phi_{e,\varepsilon}(\mathbf{u}_i)\in\mathbb{R}^{d_\text{G}}$. \\ 
         $\mathbf{C}$ & Matrix of cause embeddings as $\mathbf{C}=(\mathbf{c}_1,...,\mathbf{c}_m)$. \\ 
         $\mathbf{E}$ & Matrix of effect embeddings as $\mathbf{E}=(\mathbf{e}_1,...,\mathbf{e}_m)$. \\ 
         $\overline{\mathbf{C}}$ & Matrix of emotion cause embeddings as $\overline{\mathbf{C}}=(\overline{\mathbf{c}}_1,...,\overline{\mathbf{c}}_m)$. \\ 
         $\overline{\mathbf{E}}$ & Matrix of emotion cause embeddings as $\mathbf{E}=(\overline{\mathbf{e}}_1,...,\overline{\mathbf{e}}_m)$. \\ 
        $\boldsymbol{\mathcal{W}}$ & Trainable weights for the first biaffine module, where $\boldsymbol{\mathcal{W}}\in\mathbb{R}^{d_\text{G}\times d_\text{G}}$ \\ 
        $\boldsymbol{\mathcal{W}}_\varepsilon$ & Trainable weights for the second biaffine module, where $\boldsymbol{\mathcal{W}}_\varepsilon\in\mathbb{R}^{d_\text{G}\times |\mathcal{E}|\times d_\text{G}}$. \\ 
        \hline 
    \end{tabular}
    \caption{\label{tab:notation}Symbol notation.}
\end{table}

\end{document}